%% file: arxiv.tex
\documentclass[10pt,twocolumn,letterpaper]{article}

\usepackage{cvpr}
\usepackage{times}
\usepackage{epsfig}
\usepackage{graphicx}
\usepackage{amsmath}
\usepackage{amssymb}
\usepackage{float}
\usepackage{bbm}
\usepackage{color}
\usepackage{mathtools}
\usepackage{booktabs}       
\usepackage{tabularx}
\usepackage{microtype}      
\usepackage[symbol]{footmisc}
\usepackage[toc]{appendix}

\usepackage[labelfont=bf]{caption}
\renewcommand{\thefootnote}{\fnsymbol{footnote}}

\newcommand*\samethanks[1][\value{footnote}]{\footnotemark[#1]}
\usepackage{array}
\newcolumntype{L}[1]{>{\raggedright\arraybackslash}p{#1}}
\newcolumntype{C}[1]{>{\centering\arraybackslash}p{#1}}
\newcolumntype{R}[1]{>{\raggedleft\arraybackslash}p{#1}}

\usepackage{algorithm}
\usepackage{algpseudocode}
\algrenewcommand\algorithmicrequire{\textbf{Input:}}
\algrenewcommand\algorithmicensure{\textbf{Output:}}

\usepackage[pagebackref=true,breaklinks=true,letterpaper=true,colorlinks,bookmarks=false]{hyperref}

\cvprfinalcopy 

\def\cvprPaperID{833} 
\def\httilde{\mbox{\tt\raisebox{-.5ex}{\symbol{126}}}}

\ifcvprfinal\fi 
\begin{document}

\input{definitions}


\title{Divide and Conquer the Embedding Space for Metric Learning}

\author{
Artsiom Sanakoyeu\thanks{Both authors contributed equally to this work.} \hspace{20pt} Vadim Tschernezki\samethanks \hspace{20pt} 
Uta B{\"u}chler \hspace{20pt} Bj{\"o}rn Ommer\\
Heidelberg Collaboratory for Image Processing, IWR, Heidelberg University\\
}

\maketitle
\thispagestyle{empty} 

\begin{abstract}
Learning the embedding space, where semantically similar objects are located close together and 
dissimilar objects far apart, is a cornerstone of many computer vision applications. 
Existing approaches usually learn a single metric in the embedding space for all available data points, which may have a very complex non-uniform distribution with different notions of similarity between objects, e.g. appearance, shape, color or semantic meaning. Approaches for learning a single distance metric often struggle to encode all different types of relationships and do not  generalize well. 
In this work, we propose a novel easy-to-implement divide and conquer approach for deep metric learning, 
which significantly improves the state-of-the-art performance of metric learning.
Our approach utilizes the embedding space more efficiently by jointly splitting the embedding space and data into $K$ smaller sub-problems. 
It divides both, the data and the embedding space into $K$ subsets
and learns $K$ separate distance metrics in the non-overlapping subspaces of the embedding space, defined by groups of neurons in the embedding layer of the neural network. The proposed approach increases the convergence speed and improves generalization since the complexity of each sub-problem is reduced compared to the original one. We show that our approach outperforms the state-of-the-art by a large margin in retrieval, clustering and re-identification tasks on CUB200-2011, CARS196, Stanford Online Products, In-shop Clothes and PKU VehicleID datasets. Source code: \url{https://bit.ly/dcesml}.
\end{abstract}

\begin{figure}[t]
\begin{center}
 \includegraphics[width=\linewidth]{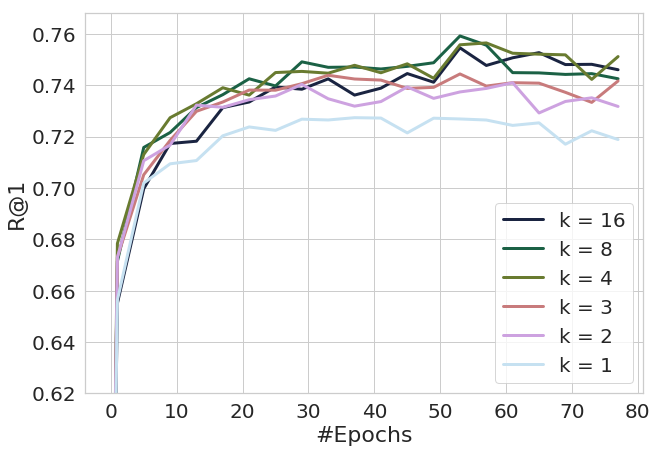} 

\end{center}
   \caption{\textbf{Evaluation of different numbers of learners}. We train our model with $K=1,2,3,4,8$ and $16$ learners on the Stanford Online Products dataset \cite{sop} and report the change of the Recall@1 score during training. An increase in the number of learners leads to higher Recall@1. The best performance is achieved with $K=8$.}
\label{fig:number_of_learners_analysis}
\end{figure}

\section{Introduction}

Deep metric learning methods learn to measure similarities or distances between arbitrary groups of data points, which is a task of paramount importance for a number of computer vision applications. Deep metric learning has been successfully applied to image search \cite{bell2015learning_contrastive,huang2016unsupervised,lifted_struct,deepranking2014}, person/vehicle re-identification \cite{chopra2005learning,vehicleid,hdc}, fine-grained retrieval \cite{proxynca}, near duplicate detection \cite{zheng2016improving}, clustering \cite{deep_clustering2016} and zero-shot learning \cite{lifted_struct,hist_loss,cliquecnn2016,sanakoyeu2018pr,buchler2018improving}.

The core idea of deep metric learning is to pull together samples with the same class label and to push apart samples coming from different classes in the learned embedding space. An embedding space with the desired properties is learned by optimizing loss functions based on pairs of images from the same or different class \cite{hadsell2006dimensionality,bell2015learning_contrastive}, triplets of images \cite{facenet,deepranking2014,tripletnet2015} or tuples of larger number of images  \cite{pddm,hist_loss,npairs,posets}, which express positive or negative relationships in the dataset.

\begin{figure*}[t]
\begin{center}
 \includegraphics[width=\linewidth]{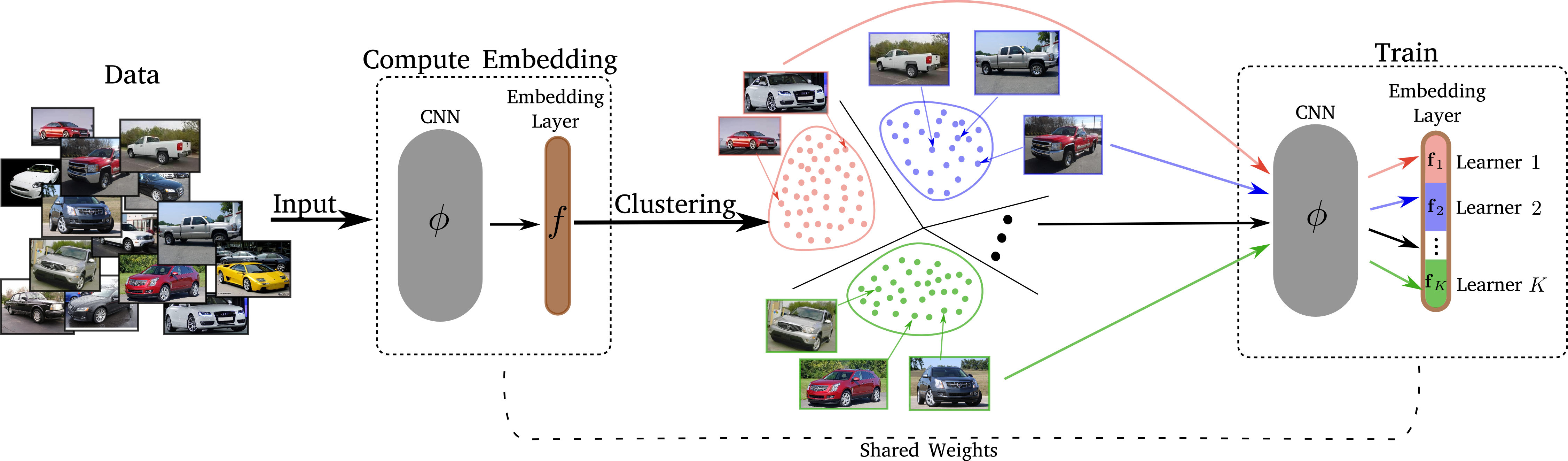} 

\end{center}
   \caption{\textbf{Pipeline of our approach.} We first cluster the data in the embedding space in $K$ groups and assign a separate subspace (learner) of the embedding layer to every cluster. During training, every learner only sees the samples assigned to the corresponding cluster.}
\label{fig:pipeline}
\end{figure*}
Existing deep metric learning approaches usually learn a single distance metric for all samples from the given data distribution. The ultimate goal for the learned metric is to resolve all conflicting relationships and pull similar images closer while pushing dissimilar images further away. However, visual data is, commonly, not uniformly distributed, but has a complex structure, where different regions of the data distribution have different densities \cite{pddm}. Data points in different regions of the distribution are often related based on different types of similarity such as shape, color, identity or semantic meaning.  While, theoretically, a deep neural network representation is powerful enough to approximate arbitrary continuous functions \cite{hornik1991approximation}, in practice this often leads to poor local minima and overfitting.
This is partially due to an inefficient usage of the embedding space \cite{lifted_struct,a_bier} and an attempt to directly fit a single distance metric to all available data \cite{rubio2015generative,li2015weakly,li2018deep}.

The problems stated above motivate an approach which will use the embedding space in a more profound way by
learning a separate distance metric for different regions of the data distribution. We propose a novel deep metric learning approach, inspired by the well-known divide and conquer algorithm. 
We explicitly split the embedding space and the data distribution into multiple parts given the network representation and learn a separate distance metric per subspace and its respective part of the distribution. Each distance metric is learned on its own subspace of the embedding space, but is based on a shared feature representation. The final embedding space is seamlessly composed by concatenating the solutions on each of the non-overlapping subspaces. See Fig.~\ref{fig:pipeline} for an illustration.

Our approach can be utilized as an efficient drop-in replacement for the final linear layer commonly used for learning embeddings in the existing deep metric learning approaches, regardless of the loss function used for training.
We demonstrate a consistent performance boost when applying our approach to the widely-used triplet loss \cite{facenet} and more complex state-of-the-art metric learning losses such as Proxy-NCA \cite{proxynca} and Margin loss \cite{margin}. 
By using the proposed approach, we achieve new state-of-the-art performance on five benchmark datasets for retrieval, clustering and re-identification: CUB200-2011 \cite{cub200_2011}, CARS196 \cite{cars196}, Stanford Online Products \cite{sop}, In-shop Clothes \cite{deepfashion}, and PKU VehicleID \cite{vehicleid}.

\section{Related work}
Metric learning has been of major interest for the vision community since its early beginnings, due to its broad applications including object retrieval \cite{lifted_struct,hist_loss,desc1}, zero-shot and single-shot learning \cite{hist_loss,lifted_struct}, keypoint descriptor learning \cite{desc_learn}, face verification \cite{chopra2005learning} and clustering \cite{deep_clustering2016}. 
With the advent of CNNs, several approaches have been proposed for supervised distance metric learning. Some methods use pairs \cite{ConvNetSimPatch} or triplets \cite{ConvNetSimTriplet,facenet} of images. Others use quadruplets \cite{npairs,hist_loss} or impose constraints on tuples of larger sizes like Lifted Structure \cite{lifted_struct}, n-pairs \cite{npairs} or poset loss \cite{posets}.

Using a tuple of images as training samples yields a huge amount of training data. However, only a small portion of the samples among all $N^p$ possible tuples of size $p$ is meaningful and provides a learning signal. 
A number of recent works tackle the problem of hard and semi-hard negative mining which provides the largest training signal by designing sampling strategies \cite{margin,smart_mining,htl,hard_triplet_gen,iscen2018mining}. 
Existing sampling techniques, however, require either running an expensive, quadratic on the number of data points preprocessing step for the entire dataset and for every epoch \cite{smart_mining,htl}, or lack global information while having a local view on the data based on a single randomly-drawn mini-batch of images \cite{facenet,margin,npairs}. 
On the contrary, our approach efficiently alleviates the problem of the abundance of easy samples, since it jointly splits the embedding space and clusters the data using the distance metric learned so far. Hence, samples inside one cluster will have smaller distances to one another than to samples from another cluster, which serves as a proxy to the mining of more meaningful relationships \cite{facenet,smart_mining}. For further details of our approach see Sec.~\ref{sec:method}.

Recently, a lot of research efforts have been devoted to designing new loss functions \cite{facility_loc,hist_loss,npairs,angular,proxynca,nca}. 
For example, Facility Location \cite{facility_loc} optimizes a cluster quality metric, Histogram loss \cite{hist_loss} minimizes the overlap between the distribution of positive and negative distances. Kihyuk Sohn proposed in \cite{npairs} the N-pairs loss which enforces a softmax cross-entropy loss among pairwise similarity values in the batch.
The Proxy-NCA loss, presented in \cite{proxynca} computes proxies for the original points in the dataset and optimizes the distances to these proxies using NCA \cite{nca}. Our work is orthogonal to these approaches and provides a framework for learning a distance metric independent on the choice of a particular loss function.

\begin{figure}[t]
\begin{center}
\includegraphics[width=\linewidth]{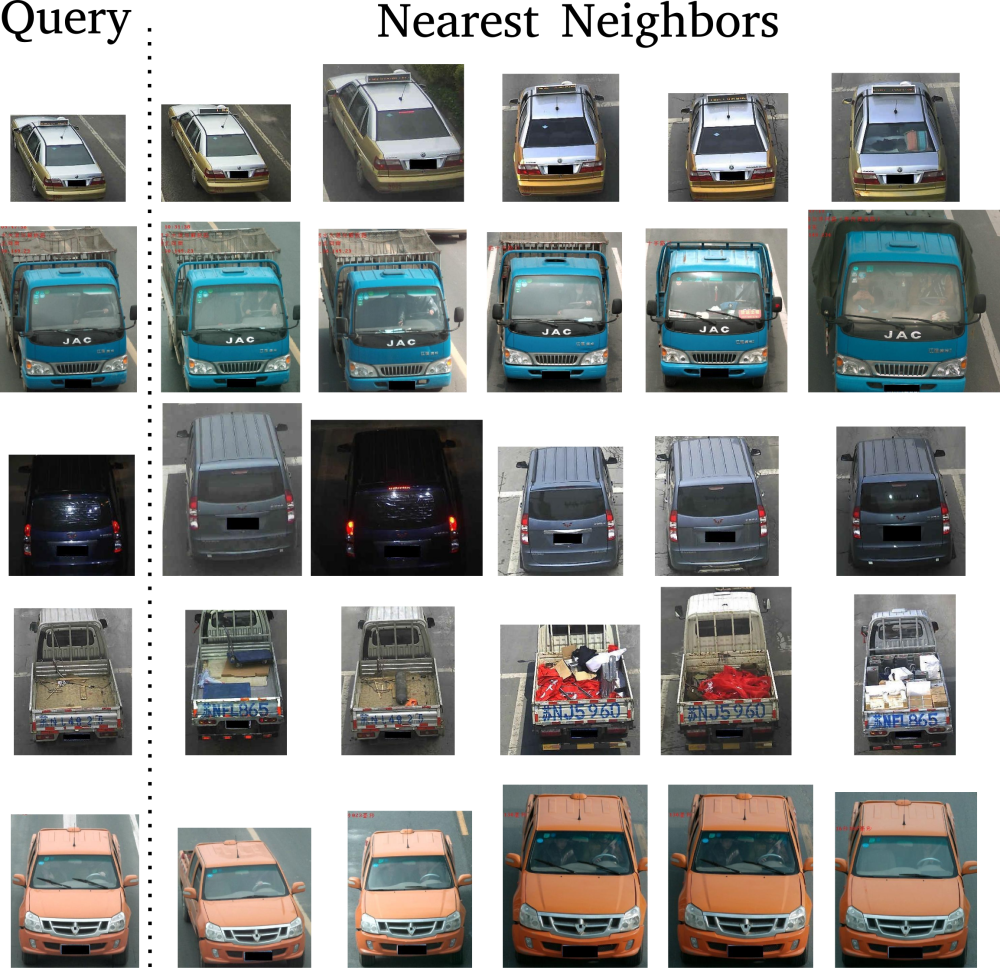} 

\end{center}
   \caption{\textbf{Qualitative image retrieval results on PKU VehicleID \cite{vehicleid}.} We show $5$ nearest neighbors per randomly chosen query image given our trained features. The queries and retrieved images are taken from the test set of the dataset.}
\label{fig:retrieval_cub}
\end{figure}

Another line of work in deep metric learning which is more related to our approach is ensemble learning \cite{hdc,a_bier,adaboost,cnn_ensemble_detection}. 
Previous works \cite{hdc,a_bier} employ a sequence of "learners" with increasing complexity and mine samples of different complexity levels for the next learners using the outputs of the previous learners. 
HDC \cite{hdc} uses a cascade of multiple models of a specific architecture, A-BIER \cite{a_bier} applies a gradient boosting learning algorithm to train several learners inside a single network in combination with an adversarial loss \cite{ganin2016domain,goodfellow2014generative}.
The key difference of the aforementioned approaches to our approach is that we split the embedding space and cluster the data jointly, so each "learner" will be assigned to the specific subspace and corresponding portion of the data. The "learners" are independently trained on non-overlapping chunks of the data which reduces the training complexity of each individual learner, facilitates the learning of decorrelated representations and can be easily parallelized. 
Moreover, our approach does not introduce extra parameters during training and works in a single model. It does not require any additional loss functions and can be applied to any existing network architecture.

\begin{figure}[t]
\begin{center}
\includegraphics[width=\linewidth]{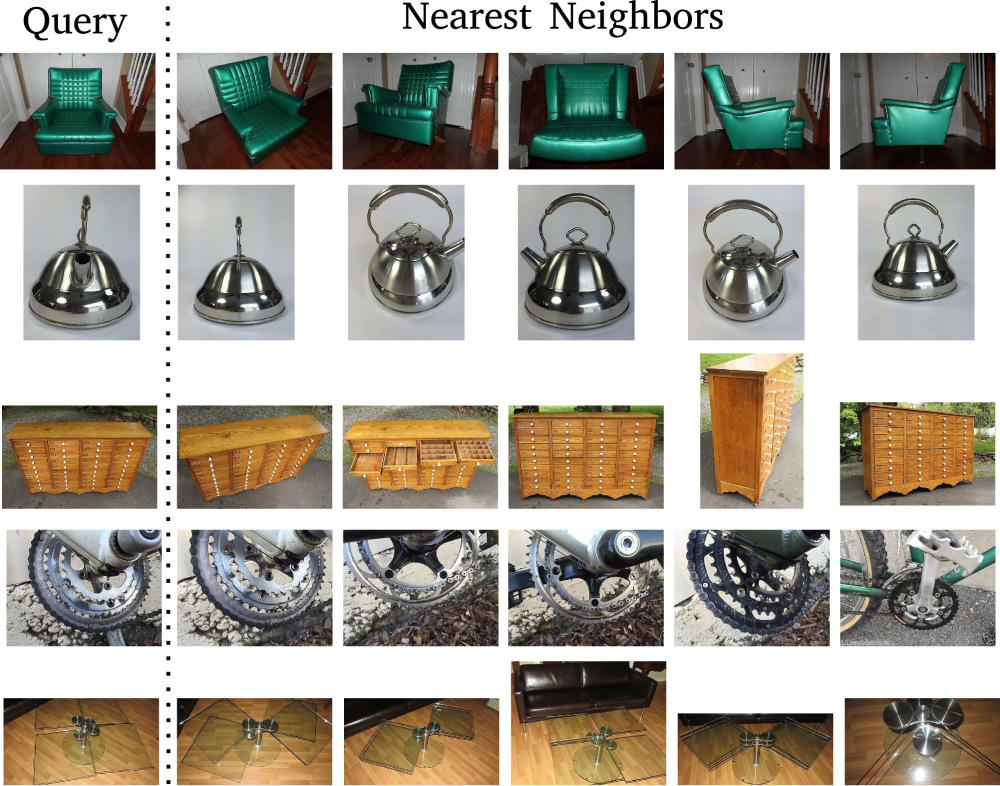}

\end{center}
   \caption{\textbf{Qualitative image retrieval results on Stanford Online Products \cite{sop}}. We randomly choose $5$ query images from the test set of the Stanford Online Products dataset and show $5$ nearest neighbors per query image given the features of our trained model. The retrieved images originate also from the test set.}
\label{fig:retrieval_sop}
\end{figure}

\section{Approach}\label{sec:method}

The main intuition behind our approach is the following: 
Solving bigger problems is usually harder than solving a set of smaller ones. 
We propose an effective and easily adaptive divide and conquer algorithm for deep metric learning.
We divide the data into multiple groups (sub-problems) to reduce the complexity
and solve the metric learning problem on each sub-problem separately.  
Since we want the data partitioning to be coupled with the current state of the embedding space, we cluster the data in the embedding space learned so far. 
Then we split the embedding layer of the network into slices. 
Each slice of the embedding layer represents an individual learner. 
Each learner is assigned to one cluster and operates in a certain 
subspace of the original embedding space. 
At the conquering stage we merge the solutions of the sub-problems, obtained by the individual learners, to get the final solution. 
We describe each step of our approach in details in Sec.~\ref{sec:method_step1} and \ref{sed:method_step2}.

\begin{figure}[t]
\begin{center}
\includegraphics[width=0.9\linewidth]{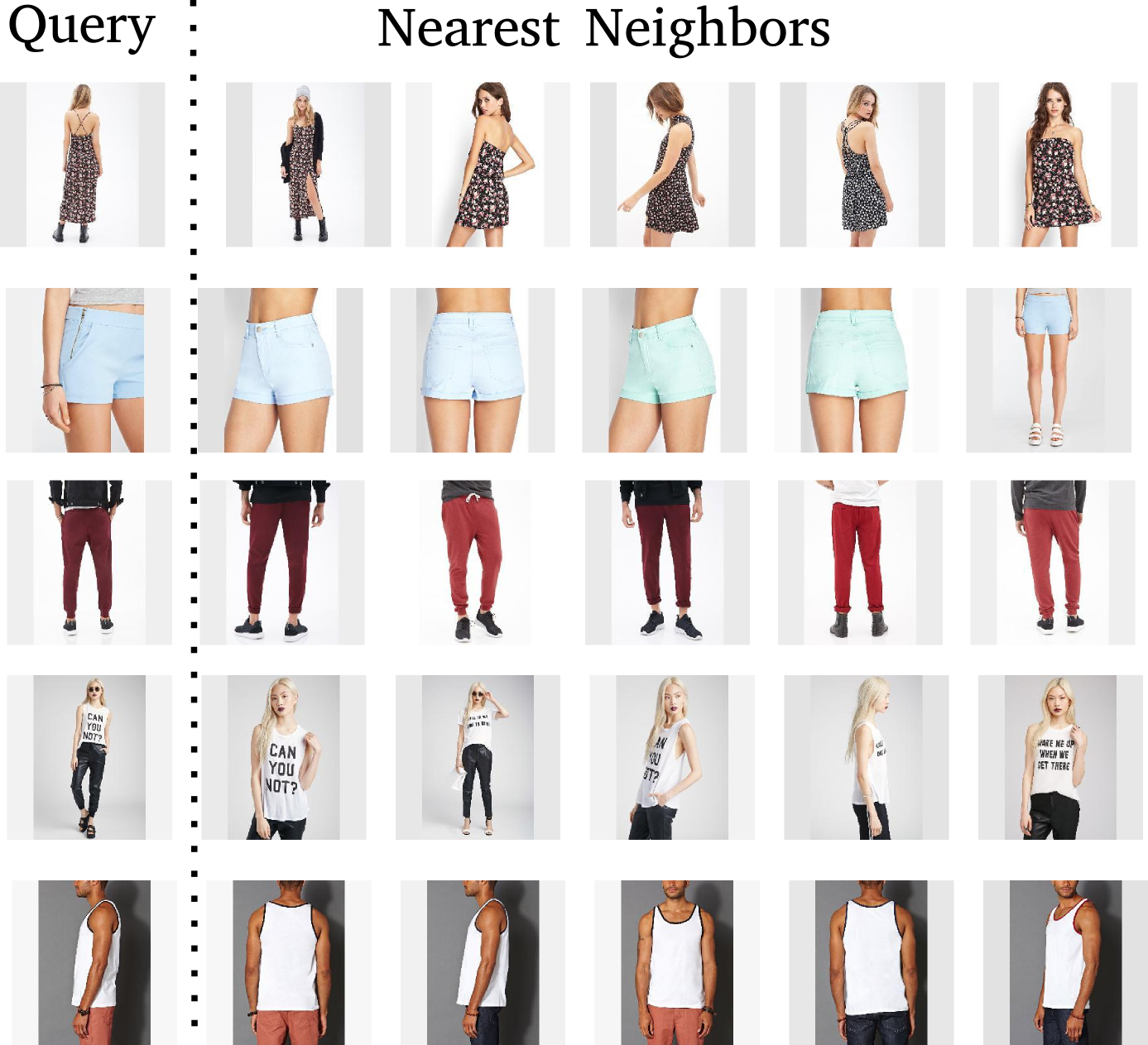} 

\end{center}
   \caption{\textbf{Qualitative image retrieval results on In-shop clothes \cite{deepfashion}.} We randomly choose $5$ query images from the query set of the In-shop clothes dataset and show $5$ nearest neighbors per query image given our trained features. The retrieved images are taken from the gallery set.}
\label{fig:retrieval_inshop}
\end{figure}

\subsection{Preliminaries} \label{sec:method_step0}
We denote the training set as $X = \{x_1, \dots, x_n\} \subset \mathcal{X} $, where $\mathcal{X}$ is the original RGB space, and the corresponding class labels as $Y = \{y_1, \dots, y_n \}$. 
A Convolutional Neural Network (CNN) learns a non-linear transformation of the image into an $m$-dimensional deep feature space $\phi(\cdot; \theta_{\phi}): \mathcal{X} \rightarrow{} \mathbb{R}^m$, where $\theta_{\phi}$ is the set of the CNN parameters. For brevity, we will use the notations $\phi(x_i; \theta_{\phi})$ and $\phi_i$ interchangeably. 

To learn a mapping into the embedding space, a linear layer $f(\cdot; \theta_f): \mathbb{R}^m \rightarrow{}  \mathbb{R}^d $ with $d$ neurons is typically appended to the CNN, where  $\theta_f$ denotes the parameters of this layer. $f(\cdot; \theta_f)$ is often normalized to have a unit length for training stability \cite{facenet}.
The goal of metric learning is to jointly learn $\phi$ and $f$ in such a way that $(f  \circ \phi)(x; \theta_{\phi}, \theta_f)$ maps similar images close to one another and dissimilar ones far apart in the embedding space. Formally, we define a distance between two data points in the embedding space as 
\begin{equation}
d_{f}(x_i, x_j) = || f(\phi_i) - f(\phi_j) ||_2.
\end{equation}
To learn the distance metric, one can use any loss function with options such as \cite{facenet,hist_loss,npairs,proxynca,margin,lifted_struct}. 
Our framework is independent of the choice of the loss function. In this paper we experiment with three different losses: Triplet loss \cite{facenet}, Proxy-NCA loss \cite{proxynca} and Margin loss \cite{margin}. For simplicity, we will demonstrate our approach in this section on the example of triplet loss, which is defined as
\begin{equation}
    l_{\text{triplet}}(a,p,n; \theta_{\phi}, \theta_{f})  = \left[ d_{f}(a, p)^2 - d_f(a, n)^2  + \alpha \right]_+, 
\end{equation}
where $[\cdot]_+$ denotes the positive part and $\alpha$ is the margin.
The triplet loss strives to keep the positive data point $p$ closer to the anchor point $a$ than any other negative point $n$. 
For brevity we omit the definitions of other losses, but we refer the interested reader to the original works \cite{proxynca,margin}.

\begin{figure}[t]
\begin{center}
\includegraphics[width=\linewidth]{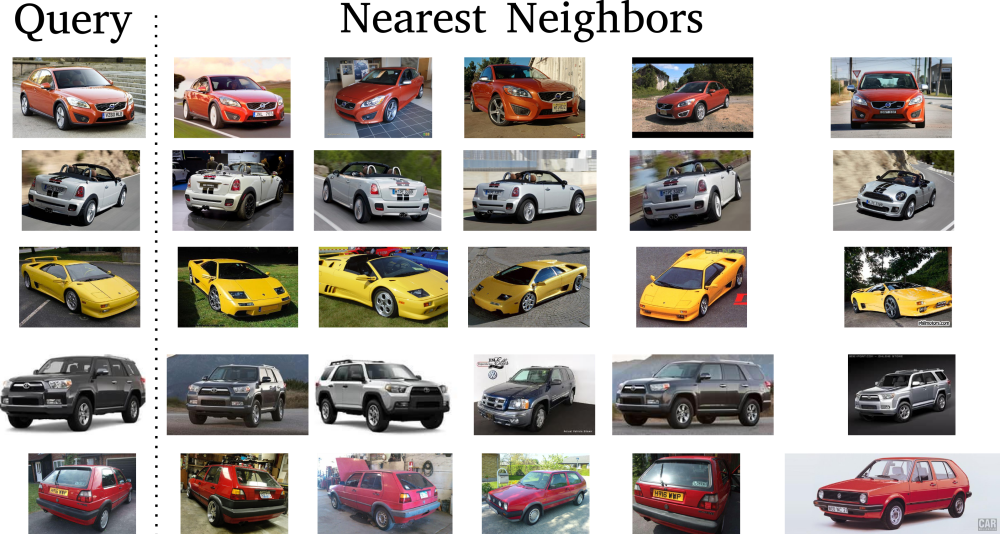} 

\end{center}
   \caption{\textbf{Qualitative image retrieval results on CARS196 \cite{cars196}.} We randomly choose $5$ query images from the test set of the CARS196 dataset and show $5$ nearest neighbors per query image given our trained features. The retrieved images are taken from the test set.}
\label{fig:retrieva_cars196}
\end{figure}

\subsection{Division of the embedding space} \label{sec:method_step1} \label{sec:division_step}
We begin with the division stage of our approach.
To reduce the complexity of the problem and to utilize the entire embedding space more efficiently we split the embedding dimensions and the data into multiple groups. Each learner will learn a separate distance metric using only a subspace of the original embedding space and a part of the data. 

\textbf{Splitting the data:} Let $K$ be the number of sub-problems. 
We group all data points $\{x_1, \dots, x_n\}$ according to their pairwise distances in the embedding space into $K$ clusters $\{C_k | 1 \le k \le K\}$ with K-means.

\textbf{Splitting the embedding:} Next, we define $K$ individual learners within the embedding space
by splitting the embedding layer of the network into $K$ consecutive slices. 
Formally, we decompose the embedding function  $f(\cdot; \theta_f)$ into $K$ functions  
$\{\f_1, \dots, \f_K\}$, where each $\f_k$ maps 
the input into the $d/K$-dimensional subspace of the original $d$-dimensional embedding space:  $\f_k(\cdot; \theta_{\f_k}): \mathbb{R}^m \rightarrow{} \mathbb{R}^{d/K}$. 
$\f_1$ will map into the first $d/K$ dimensions of the original embedding space, $\f_2$ into the second $d/K$ dimensions and so on.
Please see Fig.~\ref{fig:pipeline} for an illustration. Note that the number of the  model parameters stays constant after we perform the splitting of the embedding layer, since the learners share the underlying representation.

\subsection{Conquering stage} \label{sed:method_step2}
In this section, we first describe the step of solving individual problems. Then, we outline the merging step, where the solutions of sub-problems are combined to form the final solution.

\textbf{Training:}
After the division stage, every cluster $C_k$ is assigned to a learner $\f_k$, $1 \le k \le K$. 
Since all learners reside within a single linear embedding layer and share the underlying feature representation, we train them jointly in an alternating manner.
In each training iteration only one of the learners is updated. We uniformly sample a cluster $C_k, 1 \le k \le K$ and draw a random mini-batch $B$ from it. Then, a learner $\f_k$ minimizes its own loss defined as follows:
\begin{equation} \label{eq:learner_loss}
    \mathcal{L}_k^{\theta_{\phi}, \theta_{\f_k}} = \sum\limits_{ \mathclap{ (a,p,n) \sim B } }
    \left[ d_{\f_k}(a, p)^2 - d_{\f_k}(a, n)^2  + \alpha \right],
\end{equation}
 where triplet $(a, p, n) \in B \subset C_k$ denotes the triplets sampled from the current mini-batch, and $d_{\f_k}$ is the distance function defined in the subspace of the $k$-th learner.
As described in Eq.~\ref{eq:learner_loss} each backward pass will update only the parameters of the shared representation $\theta_{\phi}$ and the parameters of the current learner $\theta_{\f_k}$.
Motivated by the fact that the learned embedding space is improving during the time, we update the data partitioning by re-clustering every $T$ epochs using the full embedding space. The full embedding space is composed by simply concatenating the embeddings produced by the individual learners.

\textbf{Merging the solutions:}
Finally, following the divide and conquer paradigm, after individual learners converge, we merge their solutions to get the full embedding space. Merging is done by joining the embedding layer slices, corresponding to the $K$ learners, back together. 
After this, we fine-tune the embedding layer on the entire dataset 
to achieve the consistency between the embeddings of the individual learners. An overview of the full training process of our approach can be found in Algorithm~\ref{alg:training}.

\section{Experiments}
In this section, we first introduce the datasets we use for evaluating our approach and provide afterwards additional details regarding the training and testing of our framework. We then show qualitative and quantitative results which we compare with the state-of-the-art by measuring the image retrieval quality and clustering performance. The ablation study in subsection \ref{sec:abl_study} provides then some inside into our metric learning approach.

\begin{algorithm}[t]
\caption{Training a model with our approach}
\label{alg:training}
\begin{algorithmic}
\Require{$X$,$f$,$\theta_\phi$, $\theta_f$,$K$,$T$}
\Comment{\parbox[t]{.45\linewidth}{data, linear layer, CNN weights, weights of $f$,\\ \# clusters, re-cluster freq.}}\\
\Comment{cluster affiliation $\forall x_i \in X$}
\State {$\{\f_1,\dots \f_K\} \gets \textrm{SplitEmbedding}(f)$}
\Comment{set of Learners}
\State {$epoch \gets 0$}
\While{Not Converged}
\If{$epoch \text{ mod } T == 0$}
\State {$f \gets \textrm{ConcatEmbedding}(\{\f_1,\dots \f_K\})$}
\State {$emb \gets \textrm{ComputeEmbedding}(X,\theta_\phi,\theta_f$)}
\State {$\{C_1, \dots, C_K\} \gets \textrm{ClusterData}(emb,K)$}
\State {$\{\f_1,\dots \f_K\} \gets \textrm{SplitEmbedding}(f)$}
\EndIf
\Repeat
\State {$C_k \sim \{C_1, \dots, C_K\}$}
\Comment {sample cluster}
\State {$b \gets \textrm{GetBatch}(C_k)$}
\Comment {draw mini-batch}
\State{$\mathcal{L}_k \gets \textrm{FPass}(b,\theta_\phi,\theta_{\f_k})$}
\Comment {\parbox[t]{.35\linewidth}{Compute Loss of Learner $\text{\textbf{f}}_k$ (Eq. 3)}}
\State {$\theta_\phi,\theta_{\f_k} \gets \textrm{BPass}(L,\theta_\phi,\theta_{\f_k})$}
\Comment{Update weights}\\
\Until {Epoch completed}
\State {$epoch \gets epoch+1$}
\EndWhile
\State {$f \gets \textrm{ConcatEmbedding}(\{\f_1,\dots \f_K\})$}
\State {$\theta_\phi,\theta_f \gets \textrm{Finetune}(X,\theta_\phi,\theta_f,f)$}
\Ensure $\theta_\phi$, $\theta_f$
\end{algorithmic}
\end{algorithm}

\subsection{Datasets}
We evaluate the proposed approach by comparing it with the state-of-the-art on two small benchmark datasets (CARS196 \cite{cars196}, CUB200-2011 \cite{cub200_2011}), and on three large-scale datasets (Stanford Online Products \cite{sop}, In-shop Clothes \cite{deepfashion}, and PKU VehicleID \cite{vehicleid}). 
For assessing the clustering performance we utilize the normalized mutual information score \cite{schutze2008introduction}
$\text{NMI}(\Omega,\mathbb{C}) = \frac{2\cdot I(\Omega,\mathbb{C})}{H(\Omega)+H(\mathbb{C})}$, where $\Omega$ denotes the ground truth clustering and $\mathbb{C}$ the set of clusters obtained by K-means. Here $I$ represents the mutual information and $H$ the entropy. For the retrieval task we report the Recall@k metric \cite{recallk}.

\textbf{Stanford Online Products \cite{sop}} is one of the largest publicly available image collections for evaluating metric learning methods. It consists of $120,053$ images divided into $22,634$ classes of online products, where $11,318$ classes ($59,551$ images) are used for training and $11,316$ classes ($60,502$ images) for testing.  We follow the same evaluation protocol as in \cite{lifted_struct}. We calculate Recall@k score for $k=1,10,100,1000$ for evaluating the image retrieval quality and the NMI metric for appraising the clustering performance, respectively.

\textbf{CARS196} \cite{cars196} contains 196 different types of cars distributed over 16,185 images. The first 98 classes ($8,054$ images) are used for training and the other 98 classes ($8,131$ images) for testing. We train and test on the entire images without using bounding box annotations.

\textbf{CUB200-2011} \cite{cub200_2011} is an extended version of the CUB200 dataset which consolidates images of 200 different bird species with 11,788 images in total. The first 100 classes ($5,864$ images) are used for training and the second 100 classes ($5,924$ images) for testing. We train and test on the entire images without using bounding box annotations.

\begin{table}
    \centering
    \begin{tabular}{lC{0.5cm}C{0.5cm}C{0.5cm}C{0.5cm}C{0.5cm}C{0.5cm}}
    \hline
    R@k    & 1 & 10 & 100 & 1000 & NMI\\
    \hline
    Histogram \cite{hist_loss} & 63.9 & 81.7 & 92.2 & 97.7 & - \\
    Bin. Deviance \cite{hist_loss} & 65.5 & 82.3 & 92.3 & 97.6 & - \\
    Triplet Semihard \cite{facility_loc} & 66.7 & 82.4 & 91.9 & - & 89.5\\
    LiftedStruct \cite{lifted_struct} & 63.0 & 80.5 & 91.7 & 97.5 & 87.4\\
    FacilityLoc \cite{facility_loc} & 67.0 & 83.7 & 93.2 & -  & - \\
    N-pairs \cite{npairs} & 67.7 & 83.7 & 93.0 & 97.8 & 88.1\\
    Angular \cite{angular} & 70.9 & 85.0 & 93.5 & 98.0 & 88.6\\
    DAML (N-p) \cite{daml} & 68.4 & 83.5 & 92.3 & - & 89.4\\
    HDC \cite{hdc} & 69.5 & 84.4 & 92.8 & 97.7 & - \\
    DVML \cite{dvml} & 70.2 & 85.2 & 93.8 & - & 90.8\\
    BIER \cite{bier_iccv} & 72.7 & 86.5 & 94.0 & 98.0 & - \\
    ProxyNCA \cite{proxynca} & 73.7 & - & - & - & -\\
    A-BIER \cite{a_bier} & 74.2 & 86.9 & 94 & 97.8 & -\\
    HTL \cite{htl} & 74.8 & 88.3 & 94.8 & 98.4 & - \\
    Margin baseline \cite{margin} & 72.7 & 86.2 & 93.8 & 98.0 & \textbf{90.7}\\
    \textbf{Ours (Margin)} & \textbf{75.9} & \textbf{88.4} &  \textbf{94.9} & \textbf{98.1} & \underline{90.2}   \\
    \hline
    \end{tabular}
    \caption{Recall@k for $k=1,10,100,100$ and NMI on Stanford Online Products \cite{sop}}
    \label{tab:sop}
\end{table}

\textbf{In-shop Clothes Retrieval \cite{deepfashion}} contains $11,735$ classes of clothing items with $54,642$ images. We follow the evaluation protocol of \cite{deepfashion} and use a subset of  $7,986$ classes with $52,712$ images. $3,997$ classes are used for training and $3,985$ classes for testing. The test set is partitioned into query set and gallery set, containing $14,218$ and $12,612$ images, respectively. 

\textbf{PKU VehicleID \cite{vehicleid}} is a large-scale vehicle dataset that contains $221,736$ images of $26,267$ vehicles captured by surveillance cameras. The training set contains $110,178$ images of $13,134$ vehicles and the testing set comprises $111,585$ images of $13,133$ vehicles. We evaluate on $3$ test sets of different sizes as defined in \cite{vehicleid}. The small test set contains $7,332$ images of $800$ vehicles, the medium test set contains $12,995$ images of $1600$ vehicles, and the large test set contains $20,038$ images of $2400$ vehicles. This dataset has smaller intra-class variation, but it is more challenging than CARS196, because different identities of vehicles are considered as different classes, even if they share the same car model. 

\begin{table}
    \centering
    \begin{tabular}{l*{5}{C{0.06\linewidth}}}
    \hline
    R@k    & 1 & 2 & 4 & 8  & NMI\\
    \hline
    Triplet Semihard \cite{facility_loc} & 51.5 & 63.8 & 73.5 & 82.4 & 53.4 \\
    LiftedStruct \cite{lifted_struct} & 48.3 & 61.1 & 71.8 & 81.1 & 55.1\\
    FacilityLoc \cite{facility_loc} & 58.1 & 70.6 & 80.3 & 87.8 & 59.0\\
    SmartMining \cite{smart_mining} & 64.7 & 76.2 & 84.2 & 90.2 & - \\
    N-pairs \cite{npairs} & 71.1 & 79.7 & 86.5 & 91.6 & 64.0\\
    Angular \cite{angular} & 71.4 & 81.4 & 87.5 & 92.1 & 63.2 \\
    ProxyNCA \cite{proxynca} & 73.2 & 82.4 & 86.4 & 88.7 & 64.9\\
    HDC \cite{hdc} & 73.7 & 83.2 & 89.5 & 93.8 & - \\
    DAML (N-pairs) \cite{daml} & 75.1 & 83.8 & 89.7 & 93.5 & 66.0\\
    HTG \cite{hard_triplet_gen} & 76.5 & 84.7 &	90.4 & 94 & - \\
    BIER \cite{bier_iccv} & 78.0 & 85.8 & 91.1 & 95.1 & -\\
    HTL \cite{htl} & 81.4 & 88.0 & 92.7 & 95.7 & - \\
    DVML \cite{dvml} & 82.0 & 88.4 & 93.3 & 96.3 & 67.6\\
    A-BIER \cite{a_bier} & 82.0 & 89.0 & 93.2 & 96.1 & - \\
    Margin baseline \cite{margin} & 79.6 & 86.5 & 91.9 & 95.1 & 69.1\\
    \textbf{Ours (Margin)} & \textbf{84.6} & \textbf{90.7} & \textbf{94.1} & \textbf{96.5} & \textbf{70.3}\\
    \hline 
    \hline
    DREML \cite{dreml} & 86.0 & 91.7 & 95.0 & 97.2 & 76.4 \\
    \hline
    \end{tabular}
    \caption{Recall@k for $k=1,2,4,8$ and NMI on CARS196 \cite{cars196}}
    \label{tab:cars196}
\end{table}

\subsection{Implementation Details}
We implement our approach by closely following the implementation of Wu et al.~\cite{margin} based on ResNet-50 \cite{resnet}. We use an embedding of size $d=128$ and an input image size of $224 \times 224$ \cite{resnet} for all our experiments. The embedding layer is randomly initialized.
All models are trained using Adam \cite{adam} optimizer with the batch size of $80$ for Stanford Online Products and In-shop Clothes datasets, and $128$ for the other datasets. We resize the images to $256 \time 256$ and apply random crops and horizontal flips for data augmentation. For training our models we set the number of learners $K=4$ for CUB200-2011 and CARS196 due to their small size, and $K=8$ for all the other datasets. 
We have noticed that our approach is not sensitive to the values of $T$ in the range between $1$ and $10$. We set $T=2$ for all experiment, since the value alteration did not lead to significant changes in the experimental results.

Similar to \cite{margin,facenet} we initialize Margin loss with $\beta=1.2$ and Triplet loss with $\alpha=0.2$.
Mini-batches are sampled following the procedure defined in \cite{facenet,margin} with $m=4$ images per class per mini-batch for Margin loss \cite{margin} and Triplet loss \cite{facenet}, and uniformly for Proxy-NCA \cite{proxynca}.
During the clustering (Sec.~\ref{sec:division_step}) and test phase, an image embedding is composed by concatenating the embeddings of individual learners.

\begin{table}
    \centering
    \begin{tabular}{l*{5}{C{0.06\linewidth}}}
    \hline
    R@k    & 1 & 2 & 4 & 8 & NMI\\
    \hline
    LiftedStruct \cite{lifted_struct} & 46.6 & 58.1 & 69.8 & 80.2 & 56.2\\
    FacilityLoc \cite{facility_loc} & 48.2 & 61.4 & 71.8 & 81.9 & 59.2\\
    SmartMining \cite{smart_mining} & 49.8 & 62.3 & 74.1 & 83.3 & - \\
    Bin. Deviance \cite{hist_loss} & 52.8 & 64.4 & 74.7 & 83.9 & - \\
    N-pairs \cite{npairs} & 51.0 & 63.3 & 74.3 & 83.2 & 60.4\\
    DVML \cite{dvml} & 52.7 & 65.1 & 75.5 & 84.3 & 61.4 \\
    DAML (N-pairs) \cite{daml} & 52.7 & 65.4 & 75.5 & 84.3 & 61.3\\
    Histogram \cite{hist_loss} & 50.3 & 61.9 & 72.6 & 82.4 & - \\
    Angular \cite{angular} & 54.7 & 66.3 & 76.0 & 83.9 & 61.1\\
    HDC \cite{hdc} & 53.6 & 65.7 & 77.0 & 85.6 & - \\
    BIER \cite{bier_iccv} & 55.3 & 67.2 & 76.9 & 85.1 & - \\
    HTL \cite{htl} & 57.1 & 68.8 & 78.7 & 86.5 & - \\
    A-BIER \cite{a_bier} & 57.5	& 68.7 & 78.3 & 86.2 & - \\
    HTG \cite{hard_triplet_gen} & 59.5 & 71.8 & 81.3 & 88.2 & - \\
    \hline
    Triplet\textsubscript{ semihard} \cite{facility_loc} & 42.6 & 55.0 & 66.4 & 77.2 & 55.4\\
    Triplet\textsubscript{ semihard} baseline*  & 53.1 & 65.9 & 76.8 & 85.3 &	60.3 \\
    \textbf{Ours (Triplet\textsubscript{ semihard})} & 55.4 & 66.9 & 77.5 & 86.5 & 61.9 \\
    \hline 
    ProxyNCA \cite{proxynca} & 49.2 & 61.9 & 67.9 & 72.4 & 64.9 \\
    ProxyNCA baseline* & 58.7 & 70.0 & 79.1 & 87.0 & 62.5 \\
    \textbf{Ours (ProxyNCA)} & 61.8 & 73.1 & 81.8 & 88.2 & 65.7 \\
    \hline 
    Margin baseline \cite{margin} & 63.6 & 74.4 & 83.1 & 90.0 & 69.0\\
    \textbf{Ours (Margin)} & \textbf{65.9} & \textbf{76.6} & \textbf{84.4} & \textbf{90.6} & \textbf{69.6}\\
    \hline 
    \hline
    DREML \cite{dreml} & 63.9 & 75.0 & 83.1 & 89.7 & 67.8 \\
    \hline
    \end{tabular}
    \caption{Recall@k for $k=1,2,4,6,8$ and NMI on CUB200-2011 \cite{cub200_2011}. * denotes our own implementation based on ResNet-50 with $d=128$.}
    \label{tab:cub200-2011}
\end{table}

\subsection{Results}
 We now compare our approach to the state-of-the-art. 
 From Tables \ref{tab:sop}, \ref{tab:cars196}, \ref{tab:cub200-2011}, \ref{tab:inshop} and \ref{tab:vehicleid} we can see that our method with Margin loss \cite{margin} outperforms existing state-of-the-art methods on all $5$ datasets, proving its wide applicability.
 Note that we use a smaller embedding size of $d=128$ instead of $512$ employed by runner-up approaches HTL \cite{htl}, A-BIER \cite{a_bier}, BIER \cite{bier_iccv}, DVML \cite{dvml}, DAML \cite{daml}, and Angular loss \cite{angular}; HDC \cite{hdc} uses a $384$-dimensional embedding layer. Moreover, we compare our results to the deep ensembles approach DREML \cite{dreml}, which trains an ensemble of $48$ ResNet-18 \cite{resnet} networks with a total number of $537$M trainable parameters. Our model has only $25.5$M trainable parameters and still outperforms DREML \cite{dreml} on CUB200-2011 and In-shop Clothes datasets by a large margin.
 
 We demonstrate the results of our approach with three different losses on CUB200-2011: Triplet \cite{facenet}, Proxy-NCA \cite{proxynca} and Margin loss \cite{margin}. Our approach improves the Recall@1 performance by at least $2.1\%$ in each of the experiments (see Tab.~\ref{tab:cub200-2011}). This confirms that our approach is universal and can be applied to a variety of metric learning loss functions.
 We noticed that it shows especially large improvements on large-scale datasets such as on PKU VehicleID, where we improve by $3.6\%$ over the baseline with Margin loss \cite{margin} and surpass the state-of-the-art by $1\%$ in terms of Recall@1 score on the large test set. 
 We attribute this success on such a challenging dataset to the more efficient exploitation of large amounts of data due to dividing it between different learners which operate on non-overlapping subspaces of the entire embedding space.
 
 In addition to the quantitative results, we show in Figure \ref{fig:retrieval_cub},\ref{fig:retrieval_sop},\ref{fig:retrieval_inshop} and \ref{fig:retrieva_cars196} qualitative image retrieval results on CUB200-2011, Stanford Online Products, In-shop clothes, and Cars196. Note that our model is invariant to viewpoint and daylight changes.

 \begin{table}
    \centering
    \begin{tabular}{lC{0.4cm}C{0.4cm}C{0.4cm}C{0.4cm}C{0.4cm}C{0.4cm}}
    \hline
    R@k    & 1 & 10 & 20 & 30 & 50 & NMI\\
    \hline
    FashionNet \cite{deepfashion} & 53.0 & 73.0 & 76.0 & 77.0 & 80.0 & -\\
    HDC \cite{hdc} & 62.1 & 84.9 & 89.0 & 91.2 & 93.1 & - \\
    BIER \cite{bier_iccv} & 76.9 & 92.8 & 95.2 & 96.2 & 97.1 & - \\
    HTG \cite{hard_triplet_gen} & 80.3 & 93.9 & 95.8 & 96.6 & 97.1 & -\\
    HTL \cite{htl} & 80.9 & 94.3 & 95.8 & 97.2 & 97.8 & - \\
    A-BIER \cite{a_bier} & 83.1	& 95.1 & 96.9 & 97.5 & 98.0 & - \\
    Margin baseline* \cite{margin} & 82.6 &	94.8& 96.2 & 97.0 & 97.7 & 87.8 \\
    \textbf{Ours (margin)} & \textbf{85.7} & \textbf{95.5} & \textbf{96.9} & \textbf{97.5} & \textbf{98.0}
 & \textbf{88.6} \\
    \hline 
    \hline
    DREML \cite{dreml} & 78.4 & 93.7 & 95.8 & 96.7 & - & - \\
    \hline
    \end{tabular}
    \caption{Recall@k for $k=1,10,20,30,50$ and NMI on In-shop Clothes \cite{deepfashion}. * denotes our own implementation based on ResNet-50 with $d=128$.}
    \label{tab:inshop}
\end{table}

\begin{table}
    \centering
    \begin{tabular}{lC{0.4cm}C{0.4cm}|C{0.4cm}C{0.4cm}|C{0.4cm}C{0.4cm}}
    \hline
    Split Size & \multicolumn{2}{c}{Small}  & \multicolumn{2}{c}{Medium} & \multicolumn{2}{c}{Large} \\
    \hline
    R@k    & 1 & 5 & 1 & 5 & 1 & 5\\
    \hline
    Mixed Diff+CCL \cite{vehicleid} & 49.0 & 73.5 & 42.8 & 66.8 & 38.2 & 61.6 \\
    GS-TRS loss \cite{em2017incorporating} & 75.0 & 83.0 & 74.1 & 82.6 & 73.2 & 81.9\\
    BIER \cite{bier_iccv} & 82.6 & 90.6 & 79.3 & 88.3 & 76.0 & 86.4 \\
    A-BIER \cite{a_bier}  & 86.3 & 92.7 & 83.3 & 88.7 & 81.9 & 88.7 \\
    Margin baseline* \cite{margin} & 85.1 & 91.4 & 82.9 & 88.9 & 79.2 & 88.4 \\
    \textbf{Ours (margin)} & \textbf{87.7} & \textbf{92.9} & \textbf{85.7} & \textbf{90.4} & \textbf{82.9} & \textbf{90.2} \\
    \hline 
    \hline
    DREML \cite{dreml} & 88.5 & 94.8 & 87.2 & 94.2 & 83.1 & 92.4 \\
    \hline
    \end{tabular}
    \caption{Recall@k for $k=1,5$ on the small, medium and large PKU VehicleID \cite{vehicleid} dataset. * denotes our own implementation based on ResNet-50 with $d=128$.}
    \label{tab:vehicleid}
\end{table}

\subsection{Ablation Study}\label{sec:abl_study}
We perform several ablation experiments to demonstrate the effectiveness 
of the proposed method and evaluate the different components of our contribution. We use the Stanford Online Products dataset and train all models with Margin loss \cite{margin} for $80$ epochs.

First, we analyze the choice of the number of learners $K$. As can be seen in Fig.~\ref{fig:number_of_learners_analysis}, Recall@1 significantly increased already with $K = 2$. The best result is achieved with $K=8$, where each learner operates in a $16$-dimensional embedding subspace. Increasing the number of learners from $K > 1$ on, results in faster convergence and better local optima.

\begin{figure}[t]
\begin{center}
 \includegraphics[width=0.8\linewidth]{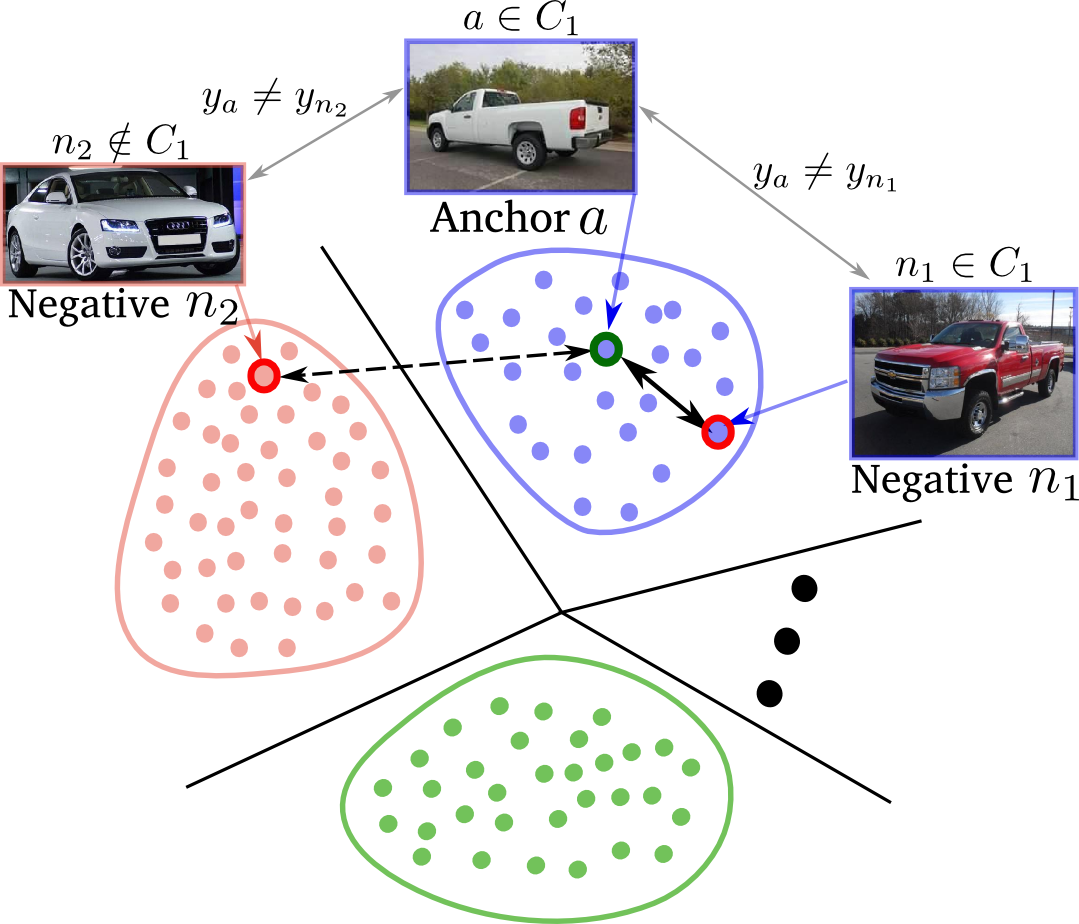} 
\end{center}
   \caption{\textbf{Natural hard negative mining.} During Training, we only sample tuples (e.g., pairs or triplets) from the same cluster. The expected value of the distance between a negative sample and an anchor within a cluster is lower than the expected value when the data points belong to different clusters. Our approach naturally finds hard negatives without explicitly performing a hard negative mining procedure.}
\label{fig:cluster_distances}
\end{figure}

\begin{figure}[t!]
\begin{center}
 \includegraphics[width=0.9\linewidth]{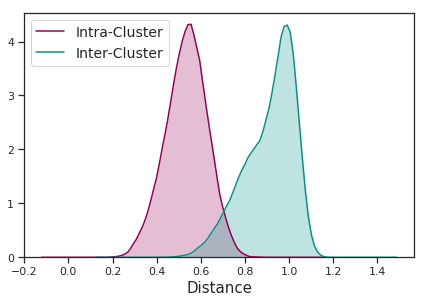} 
\end{center}
   \caption{\textbf{Intra-cluster and inter-cluster distributions of distances for negative pairs.} Red histogram shows the distribution of the pairwise distances of samples having different class labels but from the same cluster (intra-cluster); green histogram shows the distribution of the pairwise distances of samples having different class labels and drawn from different clusters (inter-cluster). Negative pairs within one cluster have lower distances and are harder on average.}
\label{fig:neg_pairs_dist_distr}
\end{figure}

Next, we study the effect of clustering the data. In Tab.~\ref{tab:clustering_evaluation} we see that substituting K-means clustering in the embedding space with random data partitioning significantly degrades the performance. On the other hand, what happens if we use K-means clustering in the embedding space, but do not split the embedding $f$ into $K$ subspaces ${\f_1, \dots, \f_K}$ during training? I.e., we perform regular training but with sampling from clusters. From Tab.~\ref{tab:clustering_evaluation} we see that it leads to a performance drop compared to the proposed approach, however it is already better than the baseline. This is due to the fact that drawing mini-batches from the clusters yields harder training samples compared to drawing mini-batches from the entire dataset. The expectation of the distance between a negative pair within the cluster is lower than the expectation of the distance between a negative pair randomly sampled from the entire dataset, as visually depicted on Fig.~\ref{fig:cluster_distances} and Fig.~\ref{fig:neg_pairs_dist_distr}.
This shows that: a) sampling from clusters provides a stronger learning signal than regular sampling from the entire dataset, b) to be able to efficiently learn from harder samples we need an individual learner for each cluster, which significantly reduces the complexity of the metric learning task. 
We also substitute K-means clustering with the fixed data partitioning, based on the ground truth labels, which are manually grouped according to semantic similarity (see "GT labels grouping" in Tab.~\ref{tab:clustering_evaluation}). We recognize that the use of a flexible clustering scheme, which depends on the data distribution in the embedding space, leads to better performance than using class labels.

\textbf{Runtime complexity:} Splitting the embedding space into subspaces and training $K$ independent learners reduces the time required for a single forward and backward pass, since we only use a $d/K$-dimensional embedding instead of the full embedding.
We perform K-means clustering every $T$ epochs. 
We use the K-means implementation from the Faiss library \cite{faiss}  which has an average complexity of $O(K n i)$, where $n$ is the number of samples, and $i$ is the number of iterations.
This adds a neglectable overhead compared to the time required for a full forward and backward pass of all images in the dataset. For example, in case of $T=2$ the clustering will add $\approx25\%$ overhead and in case of $T=8$ only $\approx6.25\%$.

\begin{table}
    \centering
    \begin{tabular}{lC{0.4cm}C{0.4cm}C{0.4cm}C{0.4cm}C{0.5cm}}
    \hline
    R@k    & 1 & 10 & 100 & 1000\\
    \hline
    Baseline \cite{margin} & 72.7 & 86.2 & 93.8 & 98.0 \\
    \shortstack[r]{K-means in the emb. space,\\ no embedding splitting} & 75.0 & 87.6 & 94.2 & 97.8\\
    \hline
    Random data partition & 73.2 & 85.8 & 93.4 & 97.6 \\
    GT labels grouping & 74.5 & 87.1 & 93.8 & 97.6 \\
    \textbf{K-means in the emb. space} & \textbf{75.9} & \textbf{88.4} &  \textbf{94.9} & \textbf{98.1} \\
    \hline
    \end{tabular}
    \caption{Evaluation of different data grouping methods on Stanford Online Products \cite{sop} with $K=8$ and Margin loss \cite{margin}.}
    \label{tab:clustering_evaluation}
\end{table}

\section{Conclusion}

We introduced a simple and efficient divide and conquer approach for deep metric learning,
which divides the data in $K$ clusters and assigns them to individual learners, constructed by splitting the network embedding layer into $K$ non-overlapping slices. We described the procedure for joint training of multiple learners within one neural network and for combining partial solutions into the final deep embedding. The proposed approach is easy to implement and can be used as an efficient drop-in replacement for a linear embedding layer commonly used in the existing deep metric learning approaches independent on the choice of the loss function.   
The experimental results on CUB200-2011 \cite{cub200_2011}, CARS196 \cite{cars196} and Stanford Online Products \cite{sop}, In-shop Clothes \cite{deepfashion} and PKU VehicleID \cite{vehicleid} show that our approach significantly outperforms the state-of-the-art on all the datasets.
\let\thefootnote\relax\footnotetext{
This work has been supported by a DFG grant OM81/1-1 and a hardware donation by NVIDIA corporation. 
}




\newpage
{\small
\bibliographystyle{ieee}
\bibliography{egbib}
}

\clearpage

\appendix
\noindent{\LARGE \textbf{Appendix}}

\section{Implementation details}
\textbf{Re-clustering every $T$ epochs:}
As pointed out in Sec.~3.3 of the main submission, we update the data partitioning by re-clustering every $T$ epochs using the full embedding space, composed by concatenating the embeddings produced by the individual learners. To maintain consistency, each learner is associated to the cluster, which is most similar to the cluster assigned to this learner in the previous iteration (i.e. in epoch $t-T$). This amounts to solving a linear assignment problem where similarity between clusters is measured in terms of IoU of points belonging to the clusters.

The source code is available at \url{https://bit.ly/dcesml}.
\section{Additional ablation study}

As discussed in the main paper, our approach facilitates the learning of decorrelated representations of individual learners. To show this, we conduct an additional ablation study where we evaluate the performance of individual learners and compute the correlation between their embeddings. In the same way as in the main paper, we use the Stanford Online Products dataset \cite{sop} and train our model with Margin loss \cite{margin}, $K=8$ and embedding size $d=128$.

\begin{figure*}[t]
\begin{center}
 \includegraphics[width=\linewidth]{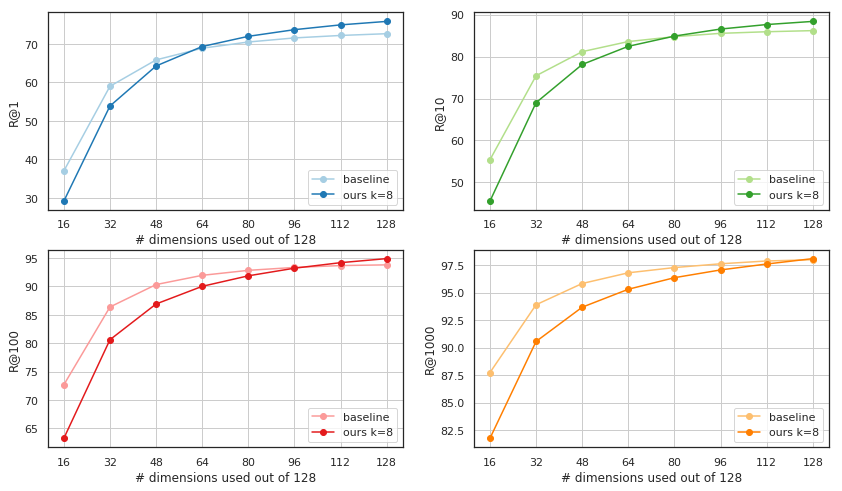} 

\end{center}
   \caption{\textbf{Evaluation of the individual learners.} We trained our model with $K=8$ learners and embedding size $d=128$ on the Stanford Online Products dataset \cite{sop}. The plots show the the Recall@k score when we use only the first $m$ out of $128$ dimensions of the embedding layer ($m=\{16, 32, \dots, 128\}$) for evaluation. Adding another $16$ dimensions corresponds to using one more learner $\f_{m/16}$ during the evaluation of our model. In case of the baseline model we do not have any learners, but for a fair comparison we also use only the first $m$ dimensions of the  embedding layer.  We see a higher performance of our approach compared to the baseline when more dimensions are used together, which shows that the individual learners in our model produce less correlated embeddings.
   }
\label{fig:number_of_emb_analysis_ablation}
\end{figure*}

We computed Recall@1 on the entire test set for every individual learner, each of which operates in a $16$-dimensional embedding subspace.
However, the baseline model was trained with only \emph{one} learner operating in the embedding space with $128$ dimensions. Hence, for comparison with the learners of our model, we split the embedding of the baseline model on $8$ non-overlapping slices of $16$ dimensions each and evaluate them separately.
In Tab.~\ref{tab:individual_learners} we can see that each individual learner trained using our approach is weaker in average than slices of the baseline model embedding. However, when we concatenate the embeddings of all individual learners together they yield significantly higher Recall@1 than the baseline model ($3.2\%$ higher in absolute values). In Fig.~\ref{fig:number_of_emb_analysis_ablation} we also show how the performance changes when we use together only $1, 2, \dots 7$ or all $8$ learners for evaluation: one learner corresponds to $16$ out of $128$ dimensions, two learners to $32$ out of $128$ dimensions and so on; $8$ learners correspond to all $128$ dimensions. We observe a larger gain compared to the baseline when more learners are used together for evaluation. This shows that the learners trained by our approach learn complementary features. 

Moreover, in Tab.~\ref{tab:individual_learners} we directly computed the correlation coefficient between the embedding produced by different learners. The correlation coefficient between the learners in our model is lower than between the slices of the baseline model embedding. 
This evidence supports our claim that the learners proposed by our approach learn less correlated features and, hence, utilize the embedding space in a more efficient way.  

\begin{figure}[t!]
\begin{center}
 \includegraphics[width=1\linewidth]{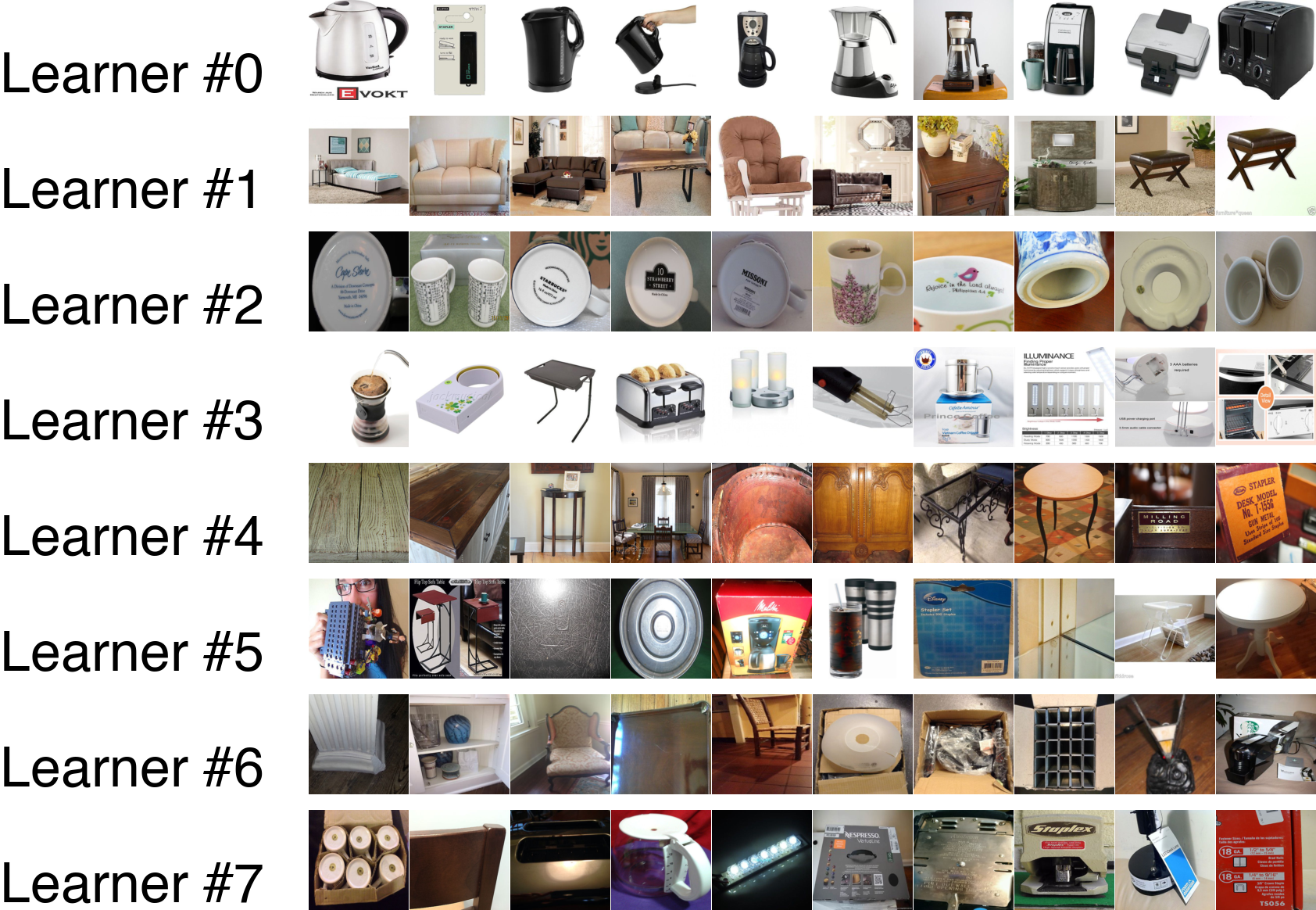}

\end{center}
  \caption{
  Representative images for the learners and their corresponding subspaces. 
    The model was trained on the Stanford Online Products dataset with $K=8$. Best viewed zoomed in.}
\label{fig:q_images}
\end{figure}

\begin{table}[t]
    \centering
    \begin{tabular}{l|cc|c}
    & Baseline & Ours & Emb. dimensions\\
    \hline
    Learner 1 & 37.0 &  29.6 & $1..16$\\
    Learner 2 & 37.0 & 29.7 &  $17..32$\\
    Learner 3 & 36.5 & 29.5 &  $33..48$\\
    Learner 4 & 36.5 & 29.4 & $49..64$ \\
    Learner 5 & 36.3 & 29.1 & $65..80$  \\
    Learner 6 & 37.4 & 29.7 & $81..96$  \\
    Learner 7 & 36.7 & 29.4 & $97..112$  \\
    Learner 8 & 37.1 & 29.9 & $113..128$  \\
    \hline
    Mean & 36.8 & 29.5 & - \\
    \hline
    \textbf{All together ($\uparrow$)} &  72.7 & \textbf{75.9} & 1..128 \\
    \hline
    \hline
    \textbf{Corr. coeff. ($\downarrow$)} & 0.0602 & \textbf{0.0498} & - \\
    \hline

    \end{tabular}
    \caption{\textbf{Evaluation of the individual learners.} Recall@1 for every individual learner on the entire test set of Stanford Online Products \cite{sop}. The last column shows the indices of the corresponding dimensions of the embedding space assigned to the learners. The individual learners of our model yield significantly higher Recall@1 than the baseline model when they are concatenated and evaluated all together, since they learn less correlated representations.}
    \label{tab:individual_learners}
\end{table}

To visualize what is captured in each embedding subspace, in Fig.~\ref{fig:q_images} we show representative images for different learners.
Every row shows $10$ query images, which are the easiest in terms of recall for one learner (R@1 $= 1$) but extremely difficult (R@30 $= 0$) for any other learner. We can see that every subspace has its own abstract "specialization". The 1st focuses on the electrical appliances, the 2nd -- on furniture, the 3rd -- on plates and mugs, etc.


\section{Additional quantitative evaluation on person re-identification}
In this section, we additionally evaluate our approach and compare to the state-of-the-art methods on Market-1501 \cite{zheng2015scalable} dataset for person re-identification.

Market-1501 \cite{zheng2015scalable} contains $32,668$ images of $1,501$ identities captured by six cameras in front of a supermarket. The $1,501$ identities are divided into a training set consisting of $12,936$ images of $751$ identities and a testing set containing the other $19,732$ images of $750$ identities. The query set contains $3,368$ images with each identity having at most 6 queries. For evaluation, we follow the standard protocol of \cite{zheng2015scalable} and report the mean average precision (mAP) and Recall@1, Recall@5 and Recall@10. In Tab.~\ref{tab:market} we demonstrate the comparison of our approach to other methods, where we can see the superior performance of the proposed approach.

\begin{table}
    \centering
    \begin{tabular}{lC{0.5cm}C{0.5cm}C{0.5cm}C{0.5cm}C{0.5cm}}
    \hline
    Recall@k   & 1 & 5 & 10 & mAP\\
    \hline
    HAP2S\_P \cite{Yu_2018_ECCV} & 84.5 & - & - & 69.7 \\
    PSE ** \cite{Sarfraz_2018_CVPR} & 87.7 & 94.5 & 96.8 & 69.0 \\ 
    HA-CNN \cite{Li_2018_CVPR} & 91.2 & - & - & 75.7 \\
    DGS \cite{Shen_2018_CVPR} & 92.7 & 96.9 & 98.1 & 82.5 \\
    DNN+CRF \cite{Chen_2018_CVPR} & 93.5 & 97.7 & - & 81.6 \\
    MGN ** \cite{Wang:2018:LDF:3240508.3240552} & 95.7 & - & - & 86.9 \\ 
    Margin baseline* \cite{margin} & 98.2 & 99.3 & 99.3 & 87.9 \\
    \textbf{Ours (Margin)} & \textbf{98.9} &  \textbf{99.5} & \textbf{99.7} & \textbf{88.8} \\
    \hline
    \end{tabular}
    \caption{Recall@k for $k=1,5,10$ and mean average precision (mAP) on Market-1501 \cite{zheng2015scalable} with single-query mode. *~denotes our own implementation based on ResNet-50 with $d=128$. ** denotes methods that use ResNet-50 as backbone.}
    \label{tab:market}
\end{table}

\end{document}

%% file: definitions.tex

\def\httilde{\mbox{\tt\raisebox{-.5ex}{\symbol{126}}}}
\def\balpha{\mbox{\boldmath $\alpha$}}
\def\bDelta{{\bf \Delta}}
\def\bLambda{{\bf \Lambda}}
\def\bvarphi{{\bf \varphi}}
\def\bTheta{{\bf \Theta}}
\def\btheta{\mbox{\boldmath $\theta$}}

\def\bPhi{\mbox{\boldmath{$\Phi$}}}
\def\vbphi{\vec{\mbox{\boldmath $\phi$}}}
\def\bb{{\bf b}}
\def\h{{\bf h}}
\def\bd{{\bf d}}
\def\ba{{\bf a}}
\def\bc{{\bf c}}
\def\p{{\bf p}}
\def\e{{\bf e}}
\def\s{{\bf s}}
\def\X{{\bf X}}
\def\x{{\bf x}}
\def\Y{{\bf Y}}
\def\y{{\bf y}}
\def\K{{\bf K}}
\def\k{{\bf k}}
\def\p{{\bf p}}
\def\bc{{\bf c}}
\def\A{{\bf A}}
\def\B{{\bf B}}
\def\C{{\bf C}}
\def\V{{\bf V}}
\def\S{{\bf S}}
\def\T{{\bf T}}
\def\W{{\bf W}}
\def\I{{\bf I}}
\def\U{{\bf U}}
\def\g{{\bf g}}
\def\G{{\bf G}}
\def\Q{{\bf Q}}
\def\d{{\bf d}}
\def\eg{{\it e.g.}}
\def\etal{{\it et. al}}
\def\H{{\bf H}}
\def\cR{{\bf R}}
\def\J{{\bf J}}
\def\bt{{\bf t}}
\def\bv{{\bf v}}
\def\R{{\bf R}}

\def\balpha{\mbox{\boldmath $\alpha$}}
\def\bdelta{\mbox{\boldmath $\delta$}}
\def\bzeta{\mbox{\boldmath $\zeta$}}
\def\bphi{\mbox{\boldmath $\phi$}}
\def\btau{\mbox{\boldmath $\tau$}}
\def\bmu{\mbox{\boldmath $\mu$}}
\def\bsigma{\mbox{\boldmath $\sigma$}}
\def\bSigma{{\bm \Sigma} }
\def\btheta{\mbox{\boldmath $\theta$}}
\def\dbphi{\dot{\mbox{\boldmath $\phi$}}}
\def\dbtau{\dot{\mbox{\boldmath $\tau$}}}
\def\dbtheta{\dot{\mbox{\boldmath $\theta$}}}
\def\bGamma{\mbox{\boldmath $\Gamma$}}
\def\bDelta{\mbox{\boldmath $\Delta$}}
\def\blambda{\mbox{\boldmath $\lambda $}}
\def\bOmega{\mbox{\boldmath $\Omega $}}
\def\bbeta{\mbox{\boldmath $\beta $}}
\def\bupsilon{\mbox{\boldmath $\Upsilon$}}
\def\myphi{\phi}
\def\bPhi{\mbox{\boldmath{$\Phi$}}}
\def\bLambda{\mbox{\boldmath{$\Lambda$}}}
\def\bSigma{\mbox{\boldmath{$\Sigma$}}}

\def\balpha{\mbox{\boldmath{$\alpha$}}}
\def\bbeta{\mbox{\boldmath{$\beta$}}}
\def\bdelta{\mbox{\boldmath{$\delta$}}}
\def\bgamma{\mbox{\boldmath{$\gamma$}}}
\def\blambda{\mbox{\boldmath{$\lambda$}}}
\def\bsigma{\mbox{\boldmath{$\sigma$}}}
\def\btheta{\mbox{\boldmath{$\theta$}}}
\def\bomega{\mbox{\boldmath{$\omega$}}}
\def\bxi{\mbox{\boldmath{$\xi$}}}

\def\bigO2{\mbox{${\cal O}$}}
\def\bigO{O}

\newcommand{\bH}{\mathbf{H}}
\def\mA{\mathcal{A}}
\def\mB{\mathcal{B}}
\def\mC{\mathcal{C}}
\def\mD{\mathcal{D}}
\def\mG{\mathcal{G}}
\def\mV{\mathcal{V}}
\def\mE{\mathcal{E}}
\def\mF{\mathcal{F}}
\def\mH{\mathcal{H}}
\def\mL{\mathcal{L}}
\def\mM{\mathcal{M}}
\def\mN{\mathcal{N}}
\def\mK{\mathcal{K}}
\def\mR{\mathcal{R}}
\def\mS{\mathcal{S}}
\def\mT{\mathcal{T}}
\def\mU{\mathcal{U}}
\def\mW{\mathcal{W}}
\def\mX{\mathcal{X}}
\def\mY{\mathcal{Y}}
\def\1n{\mathbf{1}_n}
\def\0{\mathbf{0}}
\def\1{\mathbf{1}}
\def\etal{{\em et al.}}

\def\balpha{\mbox{\boldmath $\alpha$}}
\def\bdelta{\mbox{\boldmath $\delta$}}
\def\bzeta{\mbox{\boldmath $\zeta$}}
\def\bphi{\mbox{\boldmath $\phi$}}
\def\btau{\mbox{\boldmath $\tau$}}
\def\bmu{\mbox{\boldmath $\mu$}}
\def\bsigma{\mbox{\boldmath $\sigma$}}
\def\bSigma{{\bm \Sigma} }
\def\btheta{\mbox{\boldmath $\theta$}}
\def\dbphi{\dot{\mbox{\boldmath $\phi$}}}
\def\dbtau{\dot{\mbox{\boldmath $\tau$}}}
\def\dbtheta{\dot{\mbox{\boldmath $\theta$}}}
\def\bGamma{\mbox{\boldmath $\Gamma$}}
\def\bDelta{\mbox{\boldmath $\Delta$}}
\def\blambda{\mbox{\boldmath $\lambda $}}
\def\bOmega{\mbox{\boldmath $\Omega $}}
\def\bbeta{\mbox{\boldmath $\beta $}}
\def\bupsilon{\mbox{\boldmath $\Upsilon$}}
\def\myphi{\phi}
\def\bPhi{\mbox{\boldmath{$\Phi$}}}
\def\bLambda{\mbox{\boldmath{$\Lambda$}}}
\def\bSigma{\mbox{\boldmath{$\Sigma$}}}

\def\balpha{\mbox{\boldmath{$\alpha$}}}
\def\bbeta{\mbox{\boldmath{$\beta$}}}
\def\bdelta{\mbox{\boldmath{$\delta$}}}
\def\bgamma{\mbox{\boldmath{$\gamma$}}}
\def\blambda{\mbox{\boldmath{$\lambda$}}}
\def\bsigma{\mbox{\boldmath{$\sigma$}}}
\def\btheta{\mbox{\boldmath{$\theta$}}}
\def\bomega{\mbox{\boldmath{$\omega$}}}
\def\bxi{\mbox{\boldmath{$\xi$}}}

\def\bPsi{\mbox{\boldmath $\Psi $}}
\def\bone{\mbox{\bf 1}}
\def\bzero{\mbox{\bf 0}}

\def\WB{{\bf WB}}

\def\A{{\bf A}}
\def\B{{\bf B}}
\def\C{{\bf C}}
\def\D{{\bf D}}
\def\E{{\bf E}}
\def\F{{\bf F}}
\def\G{{\bf G}}
\def\H{{\bf H}}
\def\I{{\bf I}}
\def\J{{\bf J}}
\def\K{{\bf K}}
\def\L{{\bf L}}
\def\M{{\bf M}}
\def\N{{\bf N}}
\def\O{{\bf O}}
\def\P{{\bf P}}
\def\Q{{\bf Q}}
\def\R{{\bf R}}
\def\S{{\bf S}}
\def\T{{\bf T}}
\def\U{{\bf U}}
\def\V{{\bf V}}
\def\W{{\bf W}}
\def\X{{\bf X}}
\def\Y{{\bf Y}}
\def\Z{{\bf Z}}

\def\b{{\bf b}}
\def\bc{{\bf c}}
\def\bd{{\bf d}}
\def\e{{\bf e}}
\def\f{{\bf f}}
\def\g{{\bf g}}
\def\h{{\bf h}}
\def\i{{\bf i}}
\def\j{{\bf j}}
\def\k{{\bf k}}
\def\l{{\bf l}}
\def\m{{\bf m}}
\def\n{{\bf n}}
\def\o{{\bf o}}
\def\p{{\bf p}}
\def\q{{\bf q}}
\def\br{{\bf r}}
\def\s{{\bf s}}
\def\t{{\bf t}}
\def\u{{\bf u}}
\def\v{{\bf v}}
\def\w{{\bf w}}
\def\bx{{\bf x}}
\def\y{{\bf y}}
\def\z{{\bf z}}

\def\vbphi{\vec{\mbox{\boldmath $\phi$}}}
\def\vbtau{\vec{\mbox{\boldmath $\tau$}}}
\def\vbtheta{\vec{\mbox{\boldmath $\theta$}}}
\def\vI{\vec{\bf I}}
\def\vR{\vec{\bf R}}
\def\vV{\vec{\bf V}}

\def\mvec{\vec{m}}
\def\fvec{\vec{f}}
\def\appfvec{\vec{f}_k}
\def\avec{\vec{a}}
\def\bvec{\vec{b}}
\def\evec{\vec{e}}
\def\uvec{\vec{u}}
\def\xvec{\vec{x}}
\def\wvec{\vec{w}}
\def\gradvec{\vec{\nabla}}

\def\aM{\mbox{\bf a}_M}
\def\aS{\mbox{\bf a}_S}
\def\aO{\mbox{\bf a}_O}
\def\aL{\mbox{\bf a}_L}
\def\aP{\mbox{\bf a}_P}
\def\ai{\mbox{\bf a}_i}
\def\aj{\mbox{\bf a}_j}
\def\an{\mbox{\bf a}_n}
\def\a1{\mbox{\bf a}_1}
\def\a2{\mbox{\bf a}_2}
\def\a3{\mbox{\bf a}_3}
\def\a4{\mbox{\bf a}_4}

\def\sx{\mbox{\scriptsize\bf x}}
\def\st{\mbox{\scriptsize\bf t}}
\def\ss{\mbox{\scriptsize\bf s}}
\def\cR{{\cal R}}
\def\calD{{\cal D}}
\def\calS{{\cal S}}

\def\sigmae{\sigma}
\def\sigmam{\sigma}

\def\balpha{\mbox{\boldmath{$\alpha$}}}
\def\bbeta{\mbox{\boldmath{$\beta$}}}
\def\bdelta{\mbox{\boldmath{$\delta$}}}
\def\bgamma{\mbox{\boldmath{$\gamma$}}}
\def\blambda{\mbox{\boldmath{$\lambda$}}}
\def\bsigma{\mbox{\boldmath{$\sigma$}}}
\def\btheta{\mbox{\boldmath{$\theta$}}}
\def\bomega{\mbox{\boldmath{$\omega$}}}
\def\bxi{\mbox{\boldmath{$\xi$}}}

\def\dx{{\delta \x}}
\def\dref{{\d_{ref}}}
\def\px{{\partial \x}}
\def\fxp{\f(\x, \p)}

\def\dfp{\mathbf{d}(\mathbf{f}(\x,\mathbf{p}))}
\def\dfpk{\mathbf{d}(\mathbf{f}(\x,\mathbf{p}^k))}
\def\Ep{E(\mathbf{\d, \p})}
\newcommand{\deltap}[1]{\Delta^{#1}}
\newcommand{\Jp}[1]{\J^{#1}}
\newcommand{\Hp}[1]{\H^{#1}}
\newcommand{\Hpnewton}[1]{\H^{#1}_{nt}}
\newcommand{\Psip}[1]{\Psi^{#1}}
\newcommand{\Phip}[1]{\Phi^{#1}}
\newcommand{\dPsip}[1]{\Psi_{#1}}
\newcommand{\dPhip}[1]{\Phi_{#1}}

\newcommand{\dn}{\d_{s}}
\newcommand{\dc}{\d}
\newcommand{\dnxt}[1]{\dn(\f(\x, #1))}
\newcommand{\dcur}[1]{\dc(\f(\x, #1))}

\newcommand{\one}{\mathbf{1}}
\newcommand{\zero}{\mathbf{0}}
\newcommand{\real}{\mathbb{R}}

\newcommand{\denselist}{\itemsep -1pt}
\newcommand{\sparselist}{\itemsep 1pt}